\documentclass[10pt,conference]{IEEEtran}

\usepackage[T1]{fontenc}
\usepackage[utf8]{inputenc}

\usepackage[table,dvipsnames]{xcolor}
\usepackage{paralist}
\usepackage{pifont}
\usepackage{amsmath}
\usepackage{amssymb}
\usepackage{graphicx}
\usepackage{marvosym}
\newcommand{\cmark}{\ding{51}}%
\newcommand{\xmark}{\ding{55}}%

\usepackage{newtxtext}  
\usepackage{newtxmath}  
\usepackage{inconsolata}   

\usepackage[shortlabels]{enumitem}  

\usepackage{bm}
\usepackage{booktabs}      

\usepackage{hyperref}  
\hypersetup{pdfborder={0 0 0},pagebackref=false,breaklinks=true,colorlinks=true,urlcolor=blue,linkcolor=blue,citecolor=blue,bookmarks=true}
\hypersetup{pdftitle={Transfer Learning of Keystroke Dynamics for Cross-Device User Authentication}}
\hypersetup{pdfauthor={Nuwan Kaluarachchi, Sevvandi Kandanaarachchi, Kristen Moore, Arathi Arakala, Conrad Sanderson}}

\usepackage[british]{babel}
\usepackage{hyphenat}
\usepackage[labelfont={small,bf},textfont=small]{caption}  

\usepackage{microtype}
\DisableLigatures[f]{encoding = *, family = * }  
\DisableLigatures[f,t,i,?,!,F,I,J]{encoding = *, family = * }

\def\Vec#1{{\boldsymbol{#1}}}
\def\Mat#1{{\boldsymbol{#1}}}

\begin{document}

\title{\scalebox{1.08}{\Large\bf Transfer Learning of Keystroke Dynamics for Cross-Device User Authentication}\vspace{-0.5ex}}
\author
  {
  Nuwan Kaluarachchi{\tiny~}{$^{\dagger}$},
  Sevvandi Kandanaarachchi{\tiny~}{$^{\ddagger}$},
  Kristen Moore{\tiny~}{$^{\ddagger}$},
  Arathi Arakala{\tiny~}{$^{\dagger}$},
  Conrad Sanderson{\tiny~}{$^{\ddagger\diamond}$}\\
  {\small{$^\dagger$}{\tiny~}RMIT University, Australia};
  {\small{$^\ddagger$}{\tiny~}CSIRO, Australia};
  {\small{$^\diamond$}{\tiny~}Griffith University, Australia}
  \vspace{-1ex}
  }

\maketitle

\begin{abstract}

Keystroke dynamics (typing patterns) can be used as a behavioural biometric modality for user authentication,
with applications such as fraud prevention.
While the modality has been shown to work well for single device authentication,
its application to cross-device scenarios is more challenging.
Dynamics learned on one device (e.g.,~phone)
may not be directly applicable to authentication on a secondary device with a different form factor (e.g.,~tablet)
due to changes in typing patterns that can lead to distribution drifts.
To address this,
we propose a cross-device user authentication system
based on inductive transfer learning,
where keystroke dynamics learned on one device are adapted to a secondary device.
The adapted data is then combined with necessarily limited training data for the secondary device,
which is used to robustly train a binary classifier.
Furthermore, 
an extended set of keystroke features is used to better capture discriminative dynamics.
Experiments on the BBMAS dataset show that proposed system achieves an equal error rate of~14.2\%
for the cross-device scenario,
surpassing previous methods.

\end{abstract}

\vspace{2ex}
\begin{IEEEkeywords}
transfer learning, biometrics, user authentication, cross-device, keystroke dynamics, human-computer interaction, data scarcity.
\end{IEEEkeywords}

\section{Introduction}
\vspace{1ex}

With the increasing use of passkeys for one-time password-free authentication, 
user fingerprint and face features are commonly used to ensure device security~\cite{Alrawili_2024,Cardinaux_2006}. 
However, when continuous device security is required (e.g.,~for fraud prevention in financial applications~\cite{huh2023long}),
authentication via keystroke dynamics is an alternative user-friendly approach~\mbox{\cite{shadman2025keystroke,jeong2022examining,Monrose_2000}}.

In a biometric authentication scenario, a necessarily limited number of samples of a user's biometric features is used to build a user model at registration.
When a user is required to authenticate their identity, they present their biometric sample to the device,
which compares it with the registered model and either accepts or rejects the user's identity claim.

In our increasingly connected world, users have multiple devices with varying form factors (e.g.,~phones, tablets, laptops)
and authenticate on each device separately in a manner unique to that device.
Variations in keyboard size, style and surface can cause changes in users typing patterns.
If a user were to employ continuous authentication on all their devices,
each device would need to be trained separately.
This repetition and tediousness can deter users from embracing continuous authentication,
thereby reducing overall security.

Despite the variation in keystroke dynamics due to users moving across devices,
consistency in typing styles can be exploited to transfer authentication models trained on one device to authenticate users on a secondary device~\cite{lin2022crossbehaauth}.
If enough keystroke data for each device is available, then standard machine learning models can be trained on multi-device data for cross-device authentication~\cite{belman2020doubletype}. 
However, challenges arise when 
there is limited training data for the secondary device~\cite{sun2022stratified}.

\newpage
Transfer learning problems can be categorised into three types based on their target labels~\cite{pritee2024machine,yan2024comprehensive,pan2009survey}:
\textbf{(i)}~transductive,
where the labels come from only the source domain and the labels of the target domain are predicted using the model;
\textbf{(ii)}~inductive,
where the labels are available in both the source and the target domains;
\textbf{(iii)}~unsupervised,
where no labelled data is available for both source and~target.

Transfer learning can also be categorised into two categories based on the employed feature space:
\textbf{(i)}~homogeneous and \textbf{(ii)}~heterogeneous~\cite{Bao_2026}. 
Homogeneous transfer learning indicates that the source and target feature space and label space are similar.
Heterogeneous transfer learning indicates that the source features, target features, and label spaces are different.
Lastly, transfer learning is related to domain adaptation and model transformations~\cite{Shai_2006,Sanderson_2006}.

Existing work on keystroke based authentication across devices has notable limitations.
For example, 
a temporally aware learning mechanism to authenticate users across sessions is proposed in~\cite{lin2022crossbehaauth},
where model parameters are adapted via ad-hoc fine-tuning and retraining to take into account cross-device settings,
rather than using explicit transfer learning.
Stratified transfer learning is used in~\cite{sun2022stratified},
where feature spaces from two devices are aligned into a common representation,
without explicitly learning the transformation between the~spaces. 
A~cross-domain comparison using an inductive transfer encoder to learn a transformation between domains is given in~\cite{monaco2016crossing},
where each domain uses the same device with unique settings (e.g., left- and right-hand usage),
rather than a~different device.

The three above-mentioned studies also used a small number of features (4~to~6) to represent keystroke dynamics,
potentially limiting their ability to discriminate between users.
Table~\ref{tab:litsurvey} shows salient characteristics of recent literature on keystroke dynamics using cross-device learning and/or limited target data.

In this work, we propose a novel user-specific cross-device authentication system 
based on a homogeneous inductive transfer learning approach,
which employs sufficient source device data and a limited amount of target device data to train a transfer encoder.
The source device data is then transformed via the transfer encoder to resemble target device data,
and is combined with the available target device data to form an enlarged training dataset for a binary classifier.
Furthermore, the system uses an extended set of features in order to better capture discriminative keystroke dynamics.
The proposed system is named as \textit{{T}ransfer {E}ncoder {D}ata-fusion cross-device {B}inary {C}lassification}, shortened to TEDxBC.

We continue the paper as follows.
The proposed system is detailed in Section~\ref{sec:proposed}.
The employed keystroke features are summarised in Section~\ref{sec:keystroke_features}.
Performance evaluation and an ablation study are given in Section~\ref{sec:experiments}.
The main findings and future avenues of research are given in Section~\ref{sec:conclusion}.

\begin{table}[!t]
\centering
\small
\setlength{\tabcolsep}{0.5ex}
\renewcommand{\arraystretch}{1.1}
\caption
  {
  Salient characteristics of recent literature on \mbox{keystroke} based authentication using cross-device learning (XDL) and/or limited target data (LTD).
  The text for authentication can be either restricted to fixed words/phrases, or unrestricted (free).
  }
\label{tab:litsurvey}
\begin{tabular}{lllcccc}
\textbf{Year [Ref]} & \textbf{Device} & \textbf{Text type} & \textbf{Features} &  \textbf{XDL} & \textbf{LTD} \\
\toprule
2016 \cite{monaco2016crossing}        & desktop              & fixed       &  4 & \xmark & \cmark \\
2017 \cite{cceker2017transfer}        & desktop              & free        &  2 & \xmark & \cmark \\ \midrule
2020 \cite{belman2020doubletype}      & desktop to phone     & free        &  6 & \cmark & \xmark \\
                                      & desktop to tablet    &             &    & \cmark & \xmark \\
                                      & tablet to phone      &             &    & \cmark & \xmark \\ \midrule
2022 \cite{sun2022stratified}         & phone to tablet      & fixed, free &  5 & \cmark & \cmark \\
2022 \cite{lin2022crossbehaauth}      & desktop1 to desktop2 & free        &  6 & \xmark & \cmark \\
                                      & phone to tablet      & free        &  6 & \cmark & \cmark \\ \midrule
2023 \cite{neacsu2023doublestrokenet} & desktop              & free        &  4 & \xmark & \cmark \\
                                      & phone                & free        &  4 & \xmark & \cmark \\
2024 \cite{yang2024cross}             & desktop to phone     & free        & 16 & \cmark & \xmark \\ \midrule
{this paper}                          & phone to tablet      & fixed, free & 24 & \cmark & \cmark \\
\bottomrule
\end{tabular}
\vspace{-3ex}
\end{table}

\begin{figure*}[!htb]
  \centering
  \includegraphics[width=0.79\textwidth]{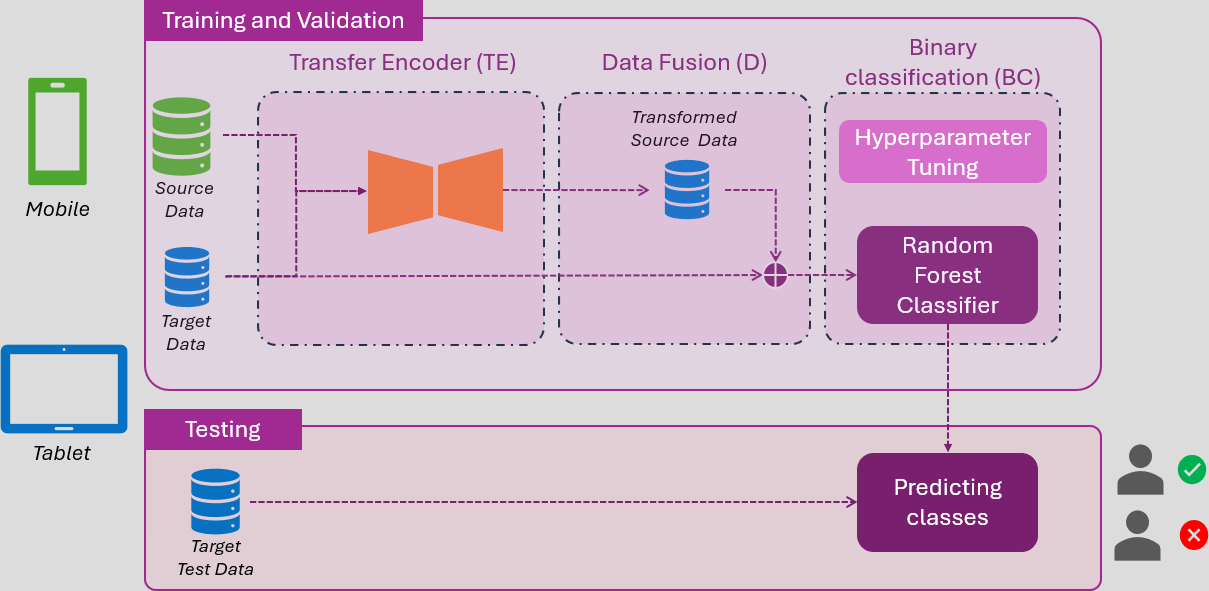}
  \caption
    {
    Overview of the proposed TEDxBC system for cross-device user authentication.
    The top section shows the training phase, comprised of 3 components:
    Transfer Encoder (TE), Data Fusion (D), and Binary Classifier (BC).
    All source data and limited target data are used for training and validation.
    The bottom section shows the testing phase, which evaluates the system's performance using unseen target~data.
    }
    \label{fig_trans_2}
    \vspace{0.4ex}
    \hrule
    \vspace{-0.4ex}
\end{figure*}

\section{Proposed Approach}
\label{sec:proposed}
\vspace{-0.5ex}

We assume the following authentication scenario.
A given user has a large amount of keystroke data from the source device (e.g.,~phone) and a limited amount of data from the target device (e.g.,~tablet).
The target device needs to be trained so that the user can be authenticated accurately on it.
However, the training process is constrained by the lack of sufficient keystroke data from the target device.
To address this challenge, we propose the TEDxBC system, summarised in Fig.~\ref{fig_trans_2}. 

The training phase is comprised of 3 components:
Transfer Encoder~(TE), Data Fusion~(D), and Binary Classifier~(BC).
The transfer encoder is a neural network which is used as a function that maps the features from the source domain to the target domain.
The data fusion component applies this function to convert the keystroke data from the source device into the target domain,
and combines it with the existing target domain data to form an enlarged training dataset for the target device.
The binary classifier then uses the augmented training dataset to build a user-specific model for the target device.
During the test phase, the trained classifier is used for determining whether a presented typing pattern belongs to the claimed user.
Each of the components is summarised in the following subsections.

\subsection{Transfer Encoder}

As labelled data is available in both the source and target domains, an inductive transfer encoder is employed~\cite{pan2009survey}.
Let \mbox{\small $X = \{\Vec{x}_i\}_{i=1}^{N}$} be a set of a user's labelled samples,
where each sample {\small $\Vec{x}_i$} is a vector of keystroke features.
The labels represent classes, where class~1 indicates samples from a given user,
while class~0 indicates impostor samples (i.e.,~samples not from to the given user).
Furthermore, let \mbox{\small $Z = \{\Vec{z}_i\}_{i=1}^{N}$} be the user's set of labelled samples (vectors) in the target domain. 

The source and target data for a user are paired using a bipartite sampling strategy~\cite{monaco2016crossing}
to create non-overlapping sets of cross-device source-target pairs for training and validation of the transfer encoder.

In brief, the \textit{encoding} function transforms the input source data
(normalised to the range [-1,+1] using min-max scaling),
into a latent space and uses a single hidden layer with the hyperbolic tangent activation function:
\begin{equation}
  \Vec{h}_i = \tanh \left( \Mat{W} \Vec{x}_i + \Vec{b}_1 \right)
\end{equation}

\noindent
where $\Mat{W}$ is a weight parameter matrix, and $\Vec{b}_1$ is a bias parameter vector.
The output latent vector $\Vec{h}_i$ has values between $-1$ and $+1$. 
Similarly, the \textit{decoding} function maps $\Vec{h}_i$ to the target domain using:
\begin{equation}
  \widehat{\Vec{z}}_i = \tanh \left( \Mat{W}^T \Vec{h}_i + \Vec{b}_2 \right)
\end{equation}

\noindent
where {\small $\widehat{\Vec{z}}_i$} is the estimated mapping for the given source data $\Vec{x}_i$ in the target domain,
{\small $\Mat{W}^T$} is known as the tied weight parameter matrix (transpose of the earlier {\small $\Mat{W}$} matrix),
and~{\small $\Vec{b}_2$} is a bias parameter vector.
We add a drop-out layer between the encoder and decoder layers, with parameter $\gamma$ indicating the drop rate.
Overall, the parameters of the transfer encoder are {\small $\left\{ \Mat{W}, \Mat{b}_1, \Mat{b}_2, \gamma \right\}$}.

In encoder-decoder architectures, it is common for the decoder's weight matrix to be chosen as the transpose of the encoder's weight matrix
for several reasons~\cite{vincent2010stacked}:
\textbf{(i)}~to~minimise the number of parameters to be learned during training,
\textbf{(ii)}~to~allow the model to capture genuine patterns using the same basis vectors used by $\Mat{W}$,
and
\textbf{(iii)}~to~generalise better to unseen data. 

The original samples from the target domain ($\Vec{z}_i$) are compared with the estimated samples $\widehat{\Vec{z}}_i$ via the loss function: 
\begin{equation}
  L \left( \Vec{z}_i,\widehat{\Vec{z}}_i  \right) = \|~ \Vec{z}_i - \widehat{\Vec{z}}_i~\|_{2}
\end{equation}

\noindent
where $\| \cdot \|_{2}$ denotes the $l^{2}$-norm and {\small $\Vec{z}_i \in Z, ~\widehat{\Vec{z}}_i \in \widehat{Z}, ~i \in [1,N]$}.

During the training process, parameters $\Mat{W}$, $\Vec{b}_1$ and $\Vec{b}_2$ are initialised using an optimal initialiser selected from a range of options,
including Glorot Uniform, Glorot Normal, He~Normal, He Uniform, and Random Uniform~\cite{Glorot_2010}.
To mitigate the risk of overfitting, we employ an early-stopping strategy.
This technique monitors the performance on the validation set specific to each user, halting training when the validation performance ceases to improve.
This approach also ensures the model generalises well to unseen data,
thereby enhancing the reliability and robustness of the user authentication system.

The loss function for each user is minimised separately,
meaning that for each user we determine the optimal initialiser, the optimal parameters, 
and the optimal number of hidden units for the encoder layers.

\subsection{Data Fusion}

As the target domain has minimal data to train a classifier, the trained and validated transfer encoder is used to transform the source data into the target domain to increase the number of samples available in the target domain.
The Data Fusion step combines this transformed source data with the limited target domain training data for a user to form an enlarged training dataset for the binary classifier.

\subsection{Binary Classifier}

The binary classifier is used to decide if a given keystroke sample belongs to the user (class~1) or not (class~0).
Random Forest was chosen as the machine learning model since it is generally considered to be robust to overfitting~\mbox{\cite{Breiman_2001,Scornet_2026}}
and has been previously used in keystroke-based authentication~\cite{sun2022stratified}.

Each user has their own classifier,
with the hyperparameters tuned to achieve optimal performance through a straightforward grid search strategy.
The search considers various parameters:
the complexity of individual trees (max\_tree\_depth),
minimum number of samples required at a leaf node (min\_samples\_leaf) and to split an internal node (min\_samples\_split).
The ensemble size (n\_estimators) is varied,
and two split quality criteria are used: Gini impurity (gini) and Information gain based on entropy (entropy). 
The list of hyperparameters and their search ranges are listed in Table~\ref{tab:hyper_classi}.

\section{Keystroke Features}
\label{sec:keystroke_features}

\vspace{1ex}

During typing on a keyboard (physical or touchscreen),
a~raw \textit{keystroke sequence} comprises the sequence of pressed keys, the direction of pressure, and the timestamps of each key press and release.
Consequently we extract 24 distinct features from such a sequence,
which are placed into three categories:
6~common temporal features such as hold times and flight times~\cite{belman2020doubletype},
16~Distance-Enhanced Flight Times features~\cite{kaluarachchi2023deft},
and 
2~so-called non-conventional features~\cite{alsultan2017non}.
The features are described in the subsections below.

\subsection{Temporal Features}

\begin{figure}[!tb]
  \centering
  \includegraphics[width=0.8\columnwidth]{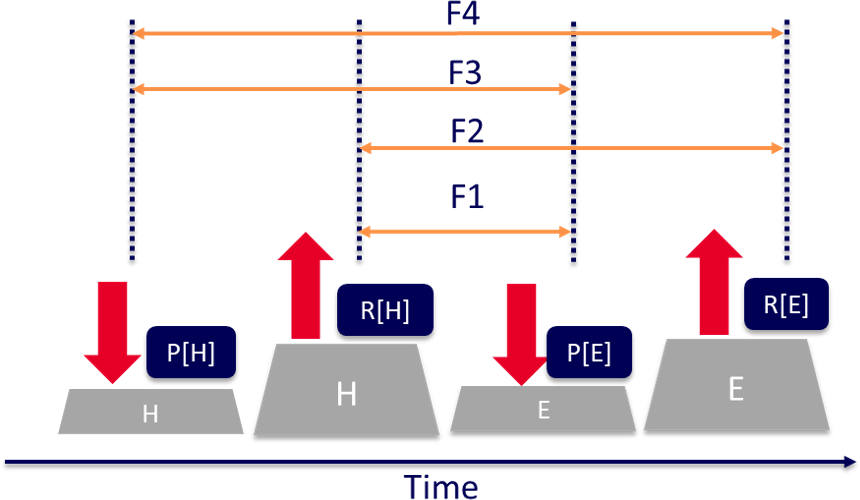}
  \caption
    {
    Press (P) and release (R) events for two keys (H and E).
    }
  \label{fig:ft}
\end{figure}

Conventional temporal features are based on press and release timing instances of single keys or adjacent key pairs in a keystroke sequence~\cite{belman2020doubletype}.
The \textit{hold time} is a feature for a single key press, calculated as the difference between the time of release and time of press of a key.
The \textit{flight time} is a feature for adjacent key pairs and based on press and release instances;
it can be further delineated into four variants~\cite{belman2020discriminative} as demonstrated in Fig.~\ref{fig:ft} and summarised below:

\begin{small}
\begin{enumerate}[{$\bullet$},leftmargin=*]
  \itemsep=1ex
  \item {Flight~1 (F1): up-down time or key interval latency}. Time between press of the second key and release of the first key.
  \item {Flight~2 (F2): up-up time or key release latency}. Time between release of the second key and release of the first key.
  \item {Flight~3 (F3): down-down time digraph time or key press latency}. Time difference between the presses of the first and second keys.
  \item {Flight~4 (F4): down-up time}. Time between release of the second key and press of the first key.    
\end{enumerate}
\end{small}

\noindent
We compute the median F1, F2, F3 and F4 values of all keypairs in a user's keystroke sequence to form the first 4 features.
We also use the median hold time of all keys in a sequence and the median hold time for a tri-graph as features.
This gives us 6 common temporal features.

\subsection{Distance-Enhanced Flight Times Features}

\begin{figure}[!tb]
  \centering
  \includegraphics[width=\columnwidth]{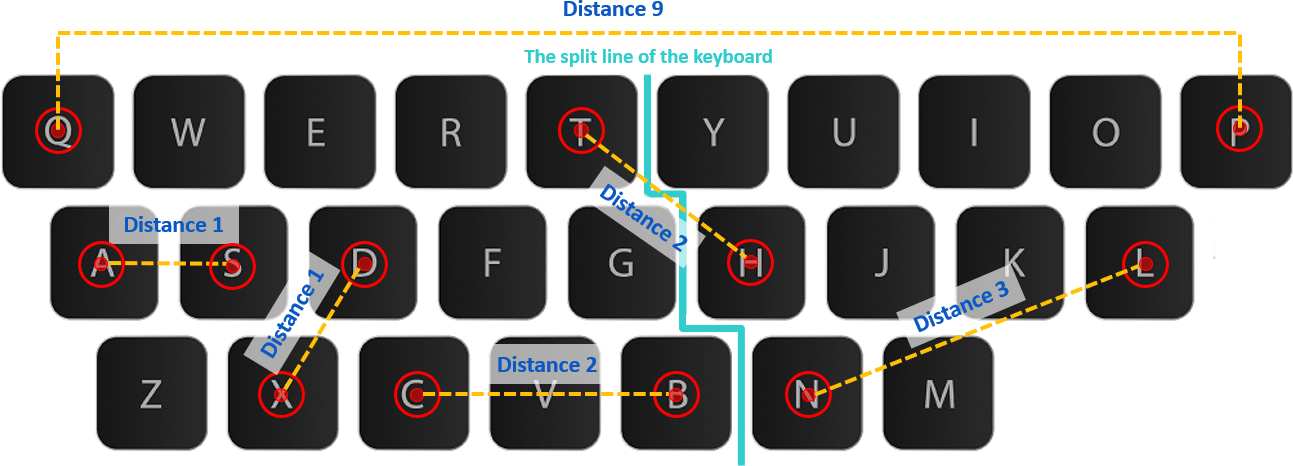}
  \caption
    {
    Calculation of distances between key pairs.
    Pair (A-S) is a distance~1 digraph,
    pair (T-H) is a distance~2 digraph,
    and
    pair (N-L) is a distance~3 digraph.
    The longest distance between a key pair is distance~9.
    The blue line separates the keyboard into left and right sides, which helps identify keys typed by the left or right hand.
    }
  \label{keyboardLayout}
  \vspace{0.4ex}
  \hrule
  \vspace{-0.4ex}
  \vspace{3ex}
\end{figure}

Distance-Enhanced Flight Times (DEFT) are temporal features defined for key pairs that are at a fixed distance from each other on a keyboard~\cite{kaluarachchi2023deft}.
Fig.~\ref{keyboardLayout} shows the QWERTY keyboard and examples of key pairs and their distances.
For example, key pairs (A-S) and (X-D) are both at a distance of~1 and are on the left side of the keyboard.
The features are defined as the median flight times of all distance \{1, 2, 3, 4\} key pairs
on either the left or right side of a keyboard, typed in a keystroke sequence.
There are a total of 32 possible DEFT features (4~flight times $\times$ 4~distances $\times$ 2~keyboard sides),
out of which we select the 16 most discriminative features via preliminary experiments~\cite{kaluarachchi2023deft}.

\subsection{Non-Conventional Features}

So-called non-conventional features are keystroke dynamic features that,
compared to the features in preceding subsections,
focus on other typing characteristics such as semi-timing and editing features as defined in~\cite{alsultan2017non}.
Based on preliminary experiments, 
we use two non-conventional features: the median error rate percentage and the median negative up-down feature.

\section{Experiments}
\label{sec:experiments}
\vspace{1ex}

\begin{table}[!t]
  \centering
  \small
  \caption{List of hyperparameters tuned via grid search for the Random Forest classifier.}
  \label{tab:hyper_classi} 
  \begin{tabular}{rl}
  \textbf{Parameter} & \textbf{Possible Values} \\
  \toprule
  max\_tree\_depth & 10, 30, 50, 70, 90, 100, None \\
  min\_samples\_leaf & 1, 2, 4 \\
  min\_samples\_split & 2, 5, 10 \\
  n\_estimators & 100, 200, 400, 600, 800, 1000 \\
  criterion & `gini', `entropy' \\
  \bottomrule
  \end{tabular}
\end{table}

We evaluate the performance of the proposed system
on the publicly available \textit{Behavioral Biometrics Multi-device and multi-Activity data from Same users} {\small (BBMAS)} dataset~\mbox{\cite{datset,belman2019insights}}.
We use data from 114 participants for which keystroke sequences are available from phones and tablets (i.e.,~a cross-device scenario).
Two types of performance evaluations are performed:
\textbf{(i)}~an ablation study,
and
\textbf{(ii)}~comparison against two existing keystroke methods that can be applied to cross-device authentication.
 
For the ablation study, 3 variants of the proposed TEDxBC system are evaluated:
\textbf{(a)}~full TEDxBC (all components);
\textbf{(b)}~TExBC, which is TEDxBC without the data fusion component (target data is not used for classifier training);
\textbf{(c)}~DxBC, which is TEDxBC without the transfer encoder component.

\newpage

For comparison against existing methods,
we use DoubleType (DT)~\cite{belman2020doubletype} and Stratified Transfer Learning (STL) \cite{sun2022stratified},
which were proposed for the cross-device paradigm and used for the phone to tablet scenario on the BBMAS dataset. 

Compared to the proposed TEDxBC approach, the DT and STL methods considerably differ in their handling of cross-device scenarios.
DT relies solely on device feature relationships without employing transfer learning techniques for cross-device adaptation.
STL aligns the source and target domain data at the class level to create a common feature space,
rather than explicitly learning the transformation between the source and target spaces. 

All methods used the same experiment setup and feature set.
The drop-out layer between the encoder and decoder layers uses an empirically determined drop rate of \mbox{$\gamma = 0.3$}.
The keystroke data from each user on each device is divided into samples.
Similar to~\cite{belman2020doubletype,sun2022stratified}, we define one sample as consisting of 150 keystrokes.
This sample size is chosen based on the typical message size limit on common social media platforms~\cite{hootsuiteIdealLength}.
With this sample size, each user has an average of 45 samples per device. 
The 45 samples for each user are partitioned into three non-overlapping subsets using a \mbox{40\%-20\%-40\%} split for both the source device (phone) and the target device (tablet). In each set, an equal number of observations that do not represent the user (feature vectors from other users) are added to create a balanced 2-class set.

In the source domain, the first two subsets are used as the training and validation sets for the transfer encoder,
while the remaining subset is reserved for transformation into the target domain.
The transformed samples are subsequently used to augment the training data for the binary classifier.
In the target domain, the first two subsets are similarly used for training and validating the transfer encoder.
The final subset is held out and used exclusively as the test set for evaluating the binary classifier.

\begin{figure}[!tb]
  \begin{minipage}[c]{0.05\columnwidth}
    \centering
    \footnotesize
    \textbf{(a)}
  \end{minipage}
  \hfill
  \begin{minipage}[c]{0.94\columnwidth}
    \centering
    \includegraphics[width=\textwidth]{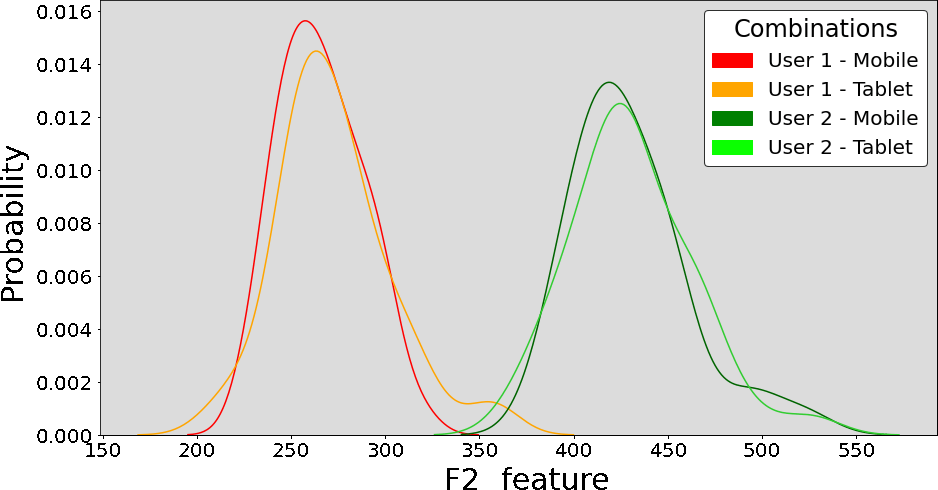}
  \end{minipage}
  
  \vspace{2ex}
  
  \begin{minipage}[c]{0.05\columnwidth}
    \centering
    \footnotesize
    \textbf{(b)}
  \end{minipage}
  \hfill
  \begin{minipage}[c]{0.94\columnwidth}
    \centering
    \includegraphics[width=\textwidth]{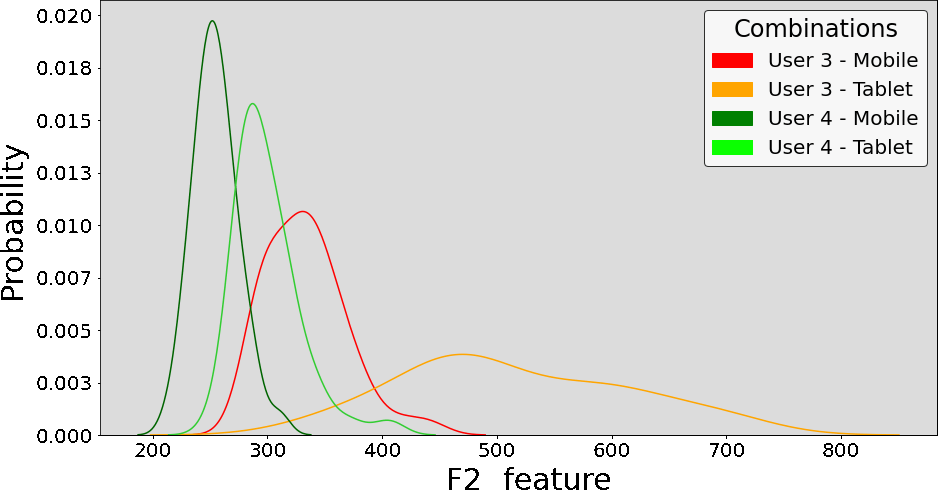}
  \end{minipage}
  \vspace{1.5ex}
  \caption
    {
    Examples of variations in the distribution of a feature across users and devices in a cross-device setting (phone and tablet).
    Panel~\textbf{(a)} shows an almost ideal situation: well separated distributions between the users, with each user having very similar distributions across two devices.
    Panel~\textbf{(b)} shows a challenging situation: overlapping distributions between the users, with each user exhibiting significant drift in the distributions across two devices.
    }
  \label{fig_motive}
  \vspace{0.4ex}
  \hrule
  \vspace{-0.4ex}
  \vspace{-2ex}
\end{figure}

\begin{figure}[!tb]
  \centering
  \includegraphics[width=0.7\columnwidth]{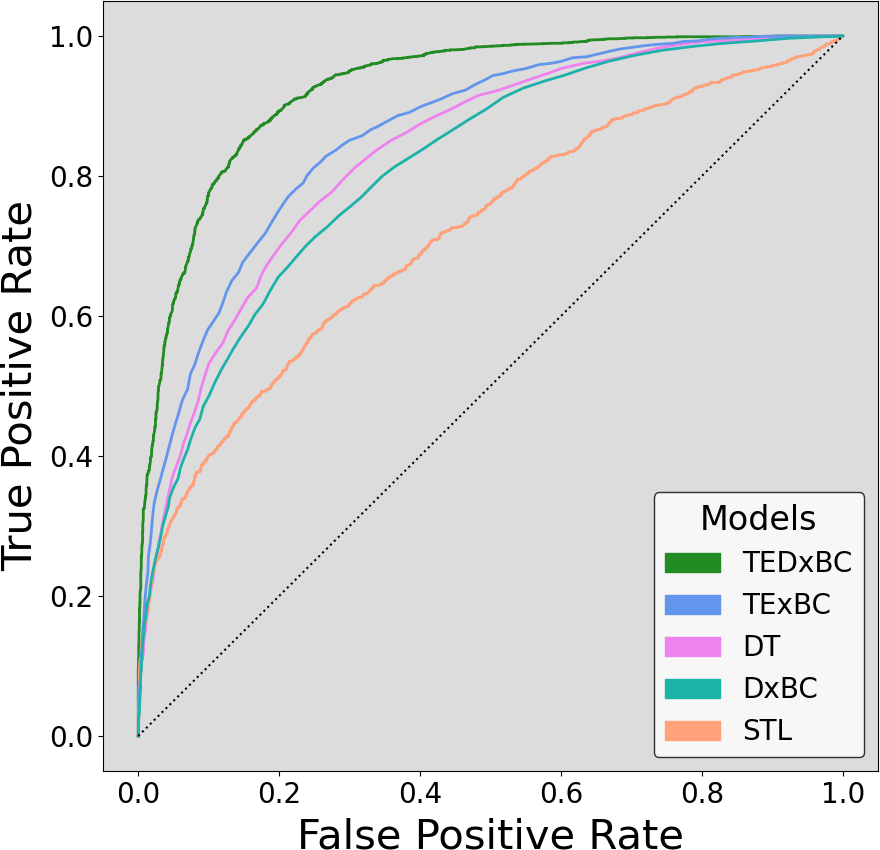}
  \caption
    {
    Performance obtained on the BBMAS dataset for:
    proposed TEDxBC method;
    DxBC~and~TExBC (ablated variants of TEDxBC);
    DoubleType (DT)~\cite{belman2020doubletype};
    Stratified Transfer Learning (STL)~\cite{sun2022stratified}.
    The closer a curve is to the top-left corner, the better the performance.
    }
  \label{fig_model_compare}
  \vspace{0.4ex}
  \vspace{-0.04ex}
  \hrule
  \vspace{-0.4ex}
  \vspace{-1ex}
\end{figure}

The following performance metrics are used: accuracy, precision, recall, F1 score and equal error rate~(EER)~\cite{Sanderson_2006,Tharwat_2020}.
For all metrics except the EER, higher values indicate better performance.
For the EER, lower values indicate better performance.
System-level metrics were computed using pooled genuine and impostor scores from all user-specific classifiers.
Examples of variations in the distribution of a feature across users and devices in a cross-device setting are shown in Fig.~\ref{fig_motive}.
The results are shown in Table~\ref{tab:model_comparision_evolution},
with the corresponding ROC curves given in Fig.~\ref{fig_model_compare}.

Overall, the proposed TEDxBC method achieves the highest performance across all metrics,
surpassing the DT and STL methods.
The ablation study indicates that both the transfer encoder and data fusion are critical components,
with the former having the most effect.
The results reveal that the transfer encoder transforms the source data well into the target domain,
and the binary classifier works effectively when it is combined with a limited amount of target data.
 
\newpage

\section{Main Findings}
\label{sec:conclusion}

In this work we have proposed a method for cross-device keystroke authentication, called TEDxBC,
suited to scenarios where the amount of training data for a secondary device is limited.
For example, a well-trained authentication model is available on a user's phone,
but authentication is required on the user's tablet,
which has a significantly different form factor that can lead to changes in keystroke dynamics.

The proposed method has three main components:
inductive transfer learning, data fusion, and user-specific binary classification.
Furthermore, an extended set of keystroke features is used to better capture discriminative keystroke dynamics of individuals.
Specifically, in addition to traditional temporal features,
we use Distance-Enhanced Flight Time features~\cite{kaluarachchi2023deft}
which include median flight times between key pairs at specific distances,
as well as non-conventional features~\cite{alsultan2017non} such as text editing events.
The design of the proposed method aims to maximise accuracy while accounting for interpretability~\cite{Sanderson_2024}.

The transfer encoder is a neural network which is used as a function that maps features from the source domain to the target domain.
The data fusion component applies this function to convert the keystroke data from the source device into the target domain,
and combines it with the existing target domain data to form an enlarged training dataset for the target device.
The binary classifier then uses the augmented training dataset to build a user-specific model for the target device.

\newpage
The transfer encoder learns the relationship between keystroke dynamics on various devices
by mapping feature distributions from one device to another.
Instead of relying solely on model fine-tuning or feature space alignment,
the proposed method dynamically captures the behavioural variations across devices,
enabling more effective and personalised cross-device authentication.
This approach ensures that user-specific typing patterns remain distinguishable even in the presence of device-induced variations,
providing a more robust and adaptable authentication mechanism.

\begin{table}[!tb]
\centering
\small
\caption
  {
  Performance obtained on the BBMAS dataset for:
  proposed TEDxBC method;
  DxBC~and~TExBC (ablated variants of TEDxBC);
  DoubleType (DT)~\cite{belman2020doubletype};
  Stratified Transfer Learning (STL)~\cite{sun2022stratified}.
  }
\label{tab:model_comparision_evolution}
\begin{tabular}{rcccccc}
\textbf{Method} & \textbf{EER} & \textbf{Accuracy} & \textbf{Precision} & \textbf{Recall} & \textbf{F1} \\
\toprule
{TEDxBC}                       & \texttt{14.2\%} & \texttt{84.1\%} & \texttt{88.6\%} & \texttt{78.9\%} & \texttt{82.8\%} \\ \midrule
TExBC                          & \texttt{22.1\%} & \texttt{78.3\%} & \texttt{78.9\%} & \texttt{69.2\%} & \texttt{72.1\%} \\ 
DxBC                           & \texttt{27.0\%} & \texttt{68.8\%} & \texttt{66.2\%} & \texttt{60.4\%} & \texttt{61.3\%} \\ \midrule
DT~\cite{belman2020doubletype} & \texttt{24.7\%} & \texttt{78.3\%} & \texttt{80.8\%} & \texttt{67.0\%} & \texttt{73.9\%} \\
STL~\cite{sun2022stratified}   & \texttt{35.0\%} & \texttt{66.8\%} & \texttt{55.9\%} & \texttt{50.1\%} & \texttt{51.0\%} \\
\bottomrule
\end{tabular}
\end{table}

Comparative evaluations on the BBMAS dataset show that the proposed method achieves an equal error rate of 14.2\% for the cross-device scenario (phone to tablet),
surpassing the performance of DoubleType~\cite{belman2020doubletype} and Stratified Transfer Learning~\cite{sun2022stratified} methods.

A limitation of the present work is that the evaluation uses a single dataset,
which may limit generalisability in real-world scenarios.
As such, in future work we aim to assess the general applicability of the proposed method
by validating its effectiveness on multiple other datasets that capture keystroke dynamics of the same individuals across various devices,
including the recent KVC-onGoing benchmark~\cite{Stragapede2025}.
It~may also be useful to evaluate the performance in more detail according to the notions of Doddington's zoo (biometric menagerie)~\cite{doddington1998sheep,mhenni2019analysis},
where authentication performance can differ based on the category of users.

\newpage

\def~{\,}  

\bibliographystyle{IEEEtran}  
\bibliography{refs}

\end{document}